\documentclass[10pt,twocolumn,letterpaper]{article}

\usepackage[pagenumbers]{cvpr} 

\usepackage{amssymb}
\usepackage{booktabs}
\usepackage{tabularx}
\usepackage{wrapfig}
\usepackage{graphicx}
\usepackage{enumitem}

\definecolor{cvprblue}{rgb}{0.21,0.49,0.74}
\usepackage[
    pagebackref,
    breaklinks,
    colorlinks,
    linkcolor=red,
    citecolor=cvprblue,
    urlcolor=cvprblue
]{hyperref}

\def\paperID{*****} 
\def\confName{CVPR}
\def\confYear{2030}

\title{Robust and Efficient Motion Reasoning for Privacy-Aware Classroom Incident Recognition}

\author{Paritosh Parmar \hspace{1cm} Landy Lan \hspace{1cm} Hong Yang \hspace{1cm} Chen Yi \hspace{1cm} Chiat Pin Tay\\
\small{Institute of High Performance Computing, Agency for Science, Technology and Research, Singapore}\\
}

\begin{document}
\maketitle

\begin{abstract}
\emph{Can computer vision help make classrooms safer?} In this pilot study, we investigate privacy-aware and computationally efficient classroom incident recognition from CCTV-style observations. This setting remains underexplored, with limited benchmarks and few methods designed for the privacy, efficiency, and generalization demands of real-world deployment.
We introduce a novel hybrid benchmark combining generative CCTV-style videos with real-world classroom pose data, and propose a lightweight, but robust motion-reasoning framework motivated by the observation that many incidents differ more in motion direction, speed, acceleration, and intensity than in pose alone. 
To that end, our method first constructs hierarchical kinematic representations of human actions. Our method then distills hierarchical, multi-order kinematic reasoning from a large teacher into a much smaller single-order student, enabling efficient per-person inference while preserving expressive motion understanding. Experiments show that our model outperforms substantially larger baselines at less than one-tenth of their computational cost, while also demonstrating stronger out-of-domain motion reasoning and zero-shot synthetic-to-real generalization. We will publicly release the benchmark, codebase, and supporting tools to facilitate further research in privacy-aware classroom safety.
\end{abstract}
\section{Introduction}

\begin{figure}
    \centering
    \includegraphics[width=0.95\linewidth]{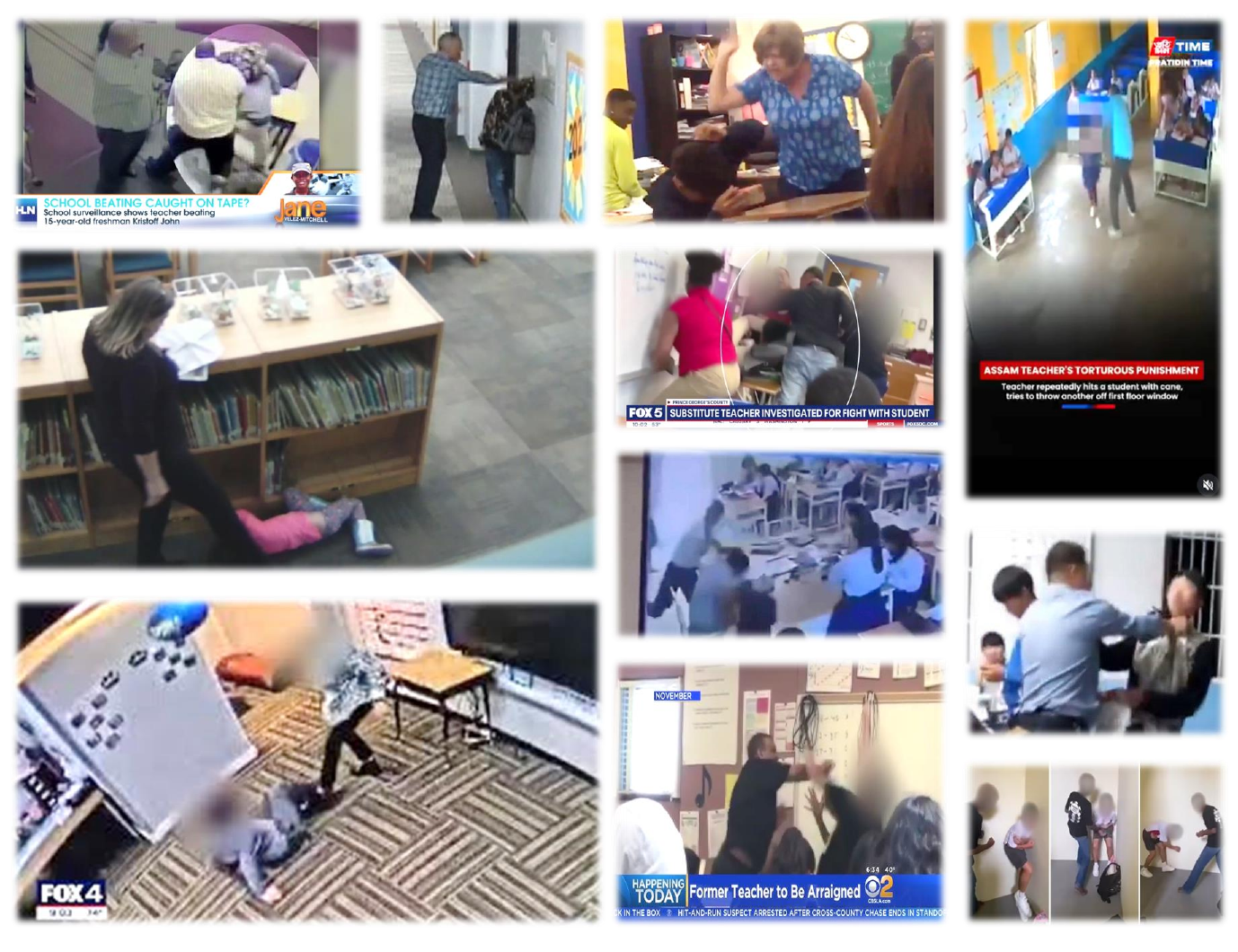}
    \caption{\textbf{Examples of classroom incidents.}}
    \label{fig:classroom_incident_examples}
\end{figure}

Schools and classrooms are environments intended to support the learning, development, and well-being of children. They are generally expected to be safe spaces in which students are protected from physical and psychological harm. However, this expectation can be violated in unfortunate circumstances. For example, students may engage in physical altercations; children may run inside classrooms, collide with furniture, and injure themselves; or, in more severe cases, students may experience physical abuse by adults (examples in \autoref{fig:classroom_incident_examples}).

Such incidents may go unnoticed or may be reported only after a delay, resulting in the loss of important contextual information and the sequence of events surrounding them. Timely access to this information can support incident review, accountability, and efforts to prevent similar events in the future.

CCTV cameras are increasingly being deployed in schools and classrooms, creating visual records of activities within these environments. However, manually monitoring continuous video streams requires sustained human attention and is inherently error-prone, making it possible for important incidents to be overlooked. Computer vision therefore has the potential to assist in automatically monitoring classroom camera feeds and identifying safety-related incidents.

Despite this potential, classroom incident recognition remains underexplored in both academic research and practical deployment. As a result, there is a lack of datasets, benchmarks, and tailored methodologies for this task. Although action recognition has been widely studied in generic settings, classroom incident recognition presents several distinct challenges.

\begin{figure}[]
    \centering
    \includegraphics[width=\linewidth]{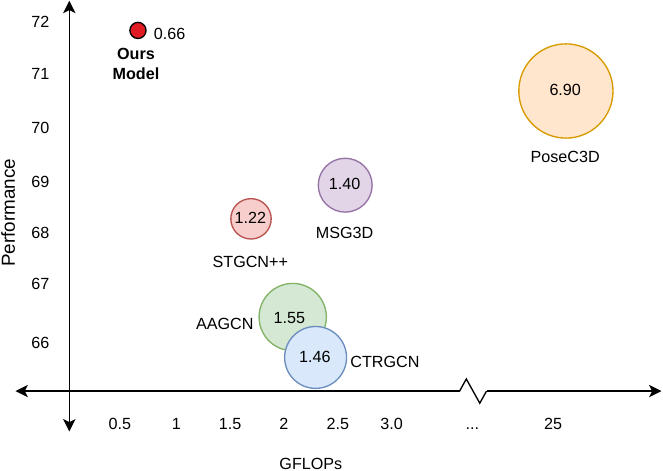}
    \caption{\textbf{Performance vs. Efficiency for various models on our synthetic dataset.} Performance \& GFLOPs are along Y \& X-axes, respectively; while circle diameter indicates the model size in million parameters (shorter diameter, smaller model size).}
    \label{fig:accuracy-efficiency}
\end{figure}

These include unusual camera viewpoints, since CCTV cameras are often mounted near classroom ceilings; the presence of both adults and young children, who are typically underrepresented in standard action recognition datasets; and domain-specific action categories, such as kicking, punching, and other safety-relevant behaviors, which may be absent from conventional benchmarks. These challenges are further compounded by privacy considerations, since classroom monitoring involves children and other identifiable individuals. This motivates recognition methods that avoid direct use of raw visual data whenever possible. In this pilot work, we aim to address these gaps. Our contributions are as follows:
\begin{itemize}[noitemsep]
    \item We introduce a CCTV-view classroom incident recognition benchmark covering safety-relevant interactions and accident scenarios involving pre-school-aged students and adult teachers. The benchmark contains synthetic videos generated using AI-based generative models along with real-world recordings.
    \item We propose a lightweight, privacy-preserving recognition framework that operates on skeletal pose features rather than raw images or video, reducing reliance on identifiable visual information.
    \item We show that the proposed method outperforms existing pose-based action-recognition baselines while requiring fewer computational resources (\autoref{fig:accuracy-efficiency}).
    \item We demonstrate that models trained on our synthetic data generalize effectively to real-world classroom footage.
\end{itemize}

\section{Related Work}

\begin{figure*}
  \centering
  \includegraphics[width=0.8\linewidth]{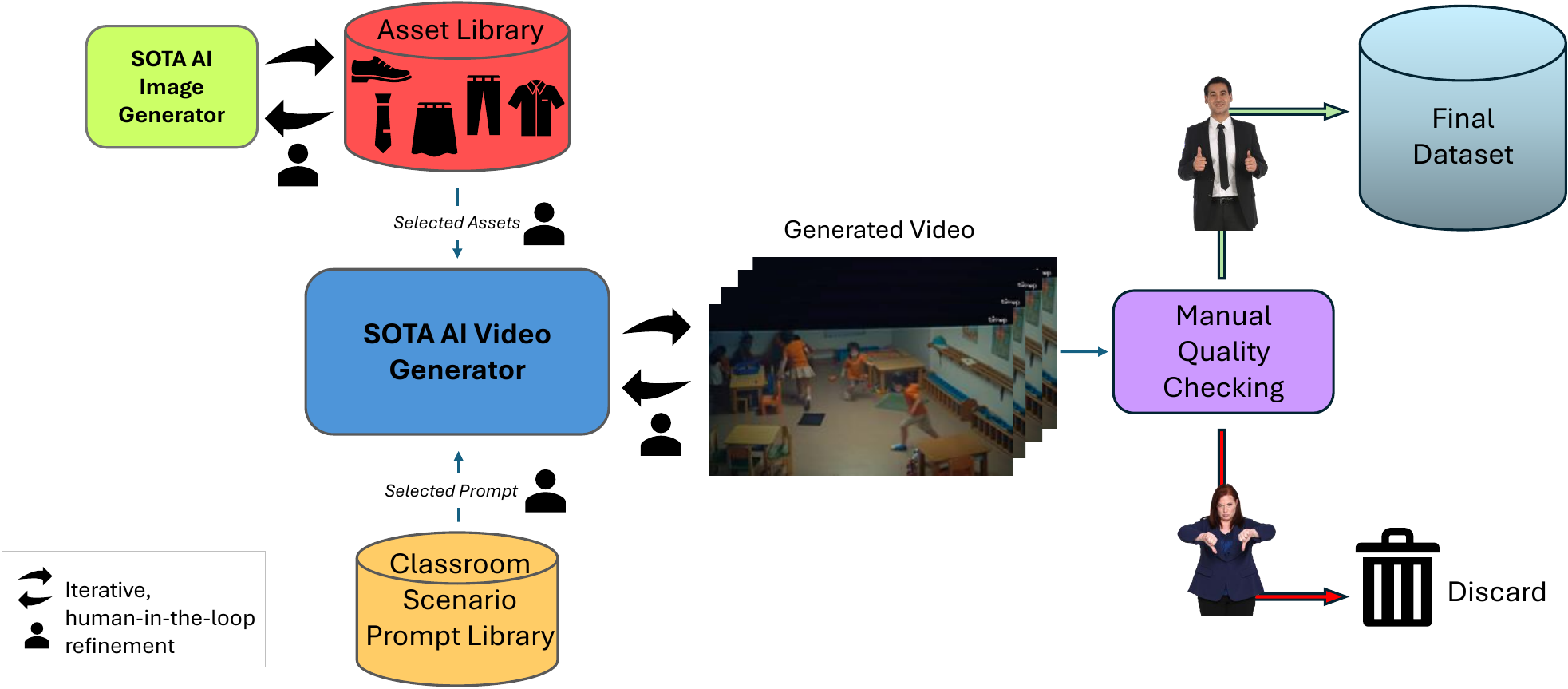}
   \caption{\textbf{Our synthetic dataset generation pipeline.}}
   \label{fig:dataset_generation_pipeline}
\end{figure*}

\noindent\textbf{Computer Vision in Education.}
Computer vision has increasingly been adopted in educational settings for classroom observation, engagement analysis, attendance tracking, and assessment~\cite{s25020373, futterer2025artificial}. Prior work has explored visual cues such as face, gaze, head pose, posture, gesture, and classroom activity to infer student attention, engagement, and instructional dynamics. Early studies and reviews  examined engagement detection in online and classroom learning, emphasizing facial expression, gaze, head pose, and posture as non-intrusive indicators of learner state~\cite{dewan2019engagement,goldberg2021attentive}. More recently, Lee and Zhai~\cite{lee2026computer} discuss the broader role of computer vision and vision-language models in STEM education research, including assessment, classroom behavior analysis, laboratory scenarios, and student-generated drawings. A recent work on handwritten answer-script evaluation combines OCR, NLP, machine learning, and deep learning to reduce examiner workload, detect marking omissions, and improve the efficiency of assessment workflows~\cite{abdullah2024design}. In STEM and active-learning contexts, researchers have proposed AI-enhanced observation systems that can automatically document student questioning, instructor lecturing, and student-led discussion, aiming to complement manual protocols such as COPUS and reduce the labor and subjectivity of human classroom observation~\cite{adeika2024transforming}. In parallel, recent work has moved toward scalable multimodal classroom analytics: Bueno \etal~\cite{bueno2026exploring} study instructional activity and discourse recognition from classroom videos and transcripts, while Tran \etal~\cite{tran2026ordinal} introduce an ordinal-aware multimodal framework for engagement recognition in collaborative learning. These works suggest that vision-based systems can support fine-grained educational analytics and teacher feedback.
Despite this progress, deploying computer vision in real educational environments remains challenging. Classroom scenes are unconstrained, with occlusion, lighting variation, changing camera viewpoints, diverse seating layouts, and context-dependent behaviors. Moreover, educational vision systems often infer sensitive states such as attention, engagement, or emotion, raising concerns about privacy, consent, surveillance, bias, and pedagogical validity. As a result, computer vision can complement classroom observation and assessment, but robust, transparent, and privacy-aware designs are necessary for responsible deployment. Existing work has largely focused on engagement, participation, discourse, assessment, or general classroom observation, leaving the identification of classroom incidents comparatively underexplored. To address this gap, we propose an efficient, privacy-preserving method for identifying classroom incidents. Specifically, we develop a first-of-its-kind dataset and a method.

\noindent\textbf{Knowledge Distillation.} The idea of transferring predictive behavior from a large model or ensemble to a smaller model predates modern knowledge distillation (KD), appearing in early work on model compression by Bucilu\u{a} \etal~\cite{buciluakdd2006} and later in shallow mimic models by Ba and Caruana~\cite{ba2014deep}. Hinton \etal~\cite{hinton2015distilling} subsequently popularized this paradigm as KD, introducing temperature-scaled soft targets to transfer ``dark knowledge'' from a teacher model to a student model. We capitalize on these advances to design an efficient student model.

\noindent\textbf{Privacy-Preserving Skeleton-Based Action Recognition.}
Skeleton-based action recognition has emerged as a privacy-conscious alternative to RGB-based video analysis, as pose sequences encode human motion while reducing reliance on the appearance, background, and identity cues present in raw video. Representative methods model skeletons as temporal sequences or spatio-temporal graphs, with graph convolutional networks such as ST-GCN learning dependencies among body joints over time~\cite{stgcn}. Subsequent methods extend this formulation through adaptive graph learning and stronger temporal modeling~\cite{shi2019two,peng2020learning}, while PoseC3D represents poses as 3D heatmap volumes to improve robustness to pose-estimation noise and cross-dataset generalization~\cite{posec3d}. Collectively, these approaches demonstrate that skeleton representations can support accurate action recognition without requiring direct use of raw video frames. However, existing privacy-aware skeleton-based methods have primarily been evaluated on general human action benchmarks, leaving their applicability to real-world educational settings comparatively underexplored. Our work addresses this gap by applying skeleton-based recognition to classroom incident identification, where privacy considerations are especially important due to the presence of minors and the sensitivity of learning environments.

\noindent\textbf{Video Generative AI.}
Video generative AI~\cite{vondrick2016generating,tulyakov2018mocogan,yan2021videogpt,hong2022cogvideo,singer2022make,ho2022imagen,villegas2022phenaki,chen2023videocrafter} aims to synthesize temporally coherent video from inputs such as text, images, audio, or existing video. Compared with SMPL-based synthesis pipelines~\cite{SMPL:2015, SMPL-X:2019, kanazawaHMR18, kocabas2019vibe}, video generative models can produce more visually realistic videos through stronger appearance-motion blending~\cite{varol2017learning, varol21_surreact}. The field has since shifted toward large-scale text-to-video and multimodal foundation models supporting image-to-video generation, reference-guided synthesis, motion control, video editing, and synchronized audio. Recent systems include Sora~\cite{openai2024sora, openai2025sora2}, Kling~\cite{klingteam2025klingomni}, Seedance~\cite{gao2025seedance}, WAN~\cite{wan2025wan}, and LTX~\cite{hacohen2025ltxvideo}. These models improve controllability through reference images, start--end frame control, video continuation, audio-video generation, and instruction-based editing. However, they still face challenges in temporal consistency, object permanence, human motion realism, physical reasoning, provenance, copyright, and misuse. In our setting, video generative AI is relevant as a tool for data augmentation and scenario simulation, but generated classroom videos must be used carefully to avoid unrealistic behavior, privacy leakage, and biased incident representations.

\noindent\textbf{Action recognition \& classroom activity understanding datasets.} Existing classroom behavior datasets~\cite{aric,bullying10k,scb_dataset,bk_sad} mainly target learning-related activities such as listening, raising hands, reading, writing, fatigue, or dozing, while general skeleton-action datasets are not tailored to classroom safety scenarios. In contrast, our dataset focuses on privacy-preserving skeleton-based classroom activity understanding with emphasis on incident-related actions such as falling, hitting, kicking, and throwing. 
Unlike mainstream action recognition datasets~\cite{liu2019ntu,shahroudy2016ntu,kay2017kinetics} built from web videos, movies, crowdsourced recordings, or controlled captures, the proposed dataset targets rare, safety-critical incidents in preschool classroom CCTV footage. It captures classroom-specific factors such as ceiling-mounted viewpoints, small human scales, teacher-child interactions, school uniforms, and incident categories including falls, fights, running, throwing, and distress-related behaviors. For further details, please see Appendix. These differences motivate a dedicated benchmark with generation and validation procedures tailored to realistic, privacy-sensitive classroom scenarios.
\section{Classroom Incident Recognition Dataset Construction}

Datasets are a crucial component of machine-learning-based systems. However, obtaining suitable data for classroom incident recognition presents several challenges, including copyright restrictions, human-subject safety, limited availability of relevant action classes, and collecting data from the target domain. For example, videos scraped from platforms such as YouTube may not be freely usable for commercial or downstream applications. Moreover, violent or safety-critical classroom actions are scarce in public data, while collecting such data through staged enactments may pose injury risks to participants.

To address these challenges, we use generative AI to produce realistic synthetic videos depicting the desired action classes in classroom settings. Although video generation has been studied for many years, recent advances in large-scale generative models have made it possible to generate high-quality and realistic videos through careful prompt design, workflow construction, and quality control. 

In the following, we present our dataset generation pipeline for creating realistic classroom videos. Whole dataset generation pipeline has been visualized in \autoref{fig:dataset_generation_pipeline}. We will release the resulting dataset to support research in classroom incident recognition. Beyond the dataset itself, the proposed pipeline provides a blueprint that researchers and practitioners can adapt to generate synthetic data for classroom understanding and related applications.

\subsection{Video Generation}

\paragraph{Selection of Action Classes.} 
We collaborated with teachers from several pre-schools in Singapore to identify dangerous classroom incidents that commonly occur and may lead to physical harm. Based on these consultations, we selected seven action classes \{{\small{\texttt{fall, punch, jump, kick, throw, run, sit}}\} for this pilot study on classroom incident recognition.
To enable models to distinguish incident classes from ordinary classroom activities, we additionally include a \texttt{background} class. This class serves as a meta-class comprising diverse classroom actions that do not fall within the selected incident categories. We hypothesize that the inclusion of a background class improves the robustness of the recognition system by reducing false positives on non-incident activities, although it also increases the difficulty of the classification task.

\paragraph{Preliminary Study on Video Generators.} 
Recent advances in video generation have led to a diverse set of generative models, each with different strengths and intended use cases. Some models are better suited for cinematic video generation, while others are optimized for stylized, animated, or short-form video content. We therefore conducted a preliminary study of several video generators, including Kling, Seedance, Sora, Veo, Wan, and LTX, to identify models suitable for generating realistic classroom videos. We evaluate these models based on visual realism, temporal consistency, controllability, action fidelity, and their ability to preserve classroom-specific visual properties. Based on this study, we selected the generators---Kling and Seedance---that best satisfied these criteria for constructing the final dataset.

\paragraph{School Uniform Asset Generation and Verification.}
Unlike many non-classroom settings, classroom videos in Singapore often involve students wearing school uniforms. To reflect this characteristic, we construct an asset library of school uniforms, with a focus on Singaporean schools. We generate school uniform and related assets using state-of-the-art generative image models in a human-in-the-loop process, where outputs are iteratively reviewed and refined to improve visual fidelity and diversity. The resulting assets cover a diverse range of school uniform styles. To ensure realism, each generated uniform asset is manually verified against the corresponding real-world uniform, and only assets that pass this verification step are included in the final library. Although our implementation focuses on Singaporean schools, our method is applicable to other schools and domains.

\paragraph{Prompt Design for Realistic Preschool CCTV Videos.}
Classrooms are typically monitored using multiple closed-circuit television (CCTV) cameras mounted near the ceiling. We worked with partner preschools and studied sample CCTV footage from their campuses to identify the desired properties of the generated videos, including typical camera placement, viewing angle, covered area, visible classroom objects, and the apparent size of students and teachers. The goal is to generate videos that resemble real CCTV footage, with students in uniforms and one or more students performing the target action. We do not limit to uniforms, and additionally generate videos with students in non-uniform clothing as well. The prompts also specify constraints such as camera distance and subject scale to ensure that human subjects appear at sizes consistent with real classroom CCTV footage.

\paragraph{Guided Video Generation \& Manual Quality Inspection.} 
We provide the generated prompts and uniform assets to the video generator to conditionally synthesize classroom videos. Each generated video is then independently reviewed by three authors. A video is retained only if all reviewers agree that it satisfies the specified requirements. Specifically, reviewers assess: 1) the visual realism of human subjects and objects, including their temporal consistency; 2) camera-related properties, such as viewpoint, angle, and subject scale; 3) motion realism; and 4) action semantics.  

\begin{figure*}[t]
  \centering
  \includegraphics[width=\linewidth]{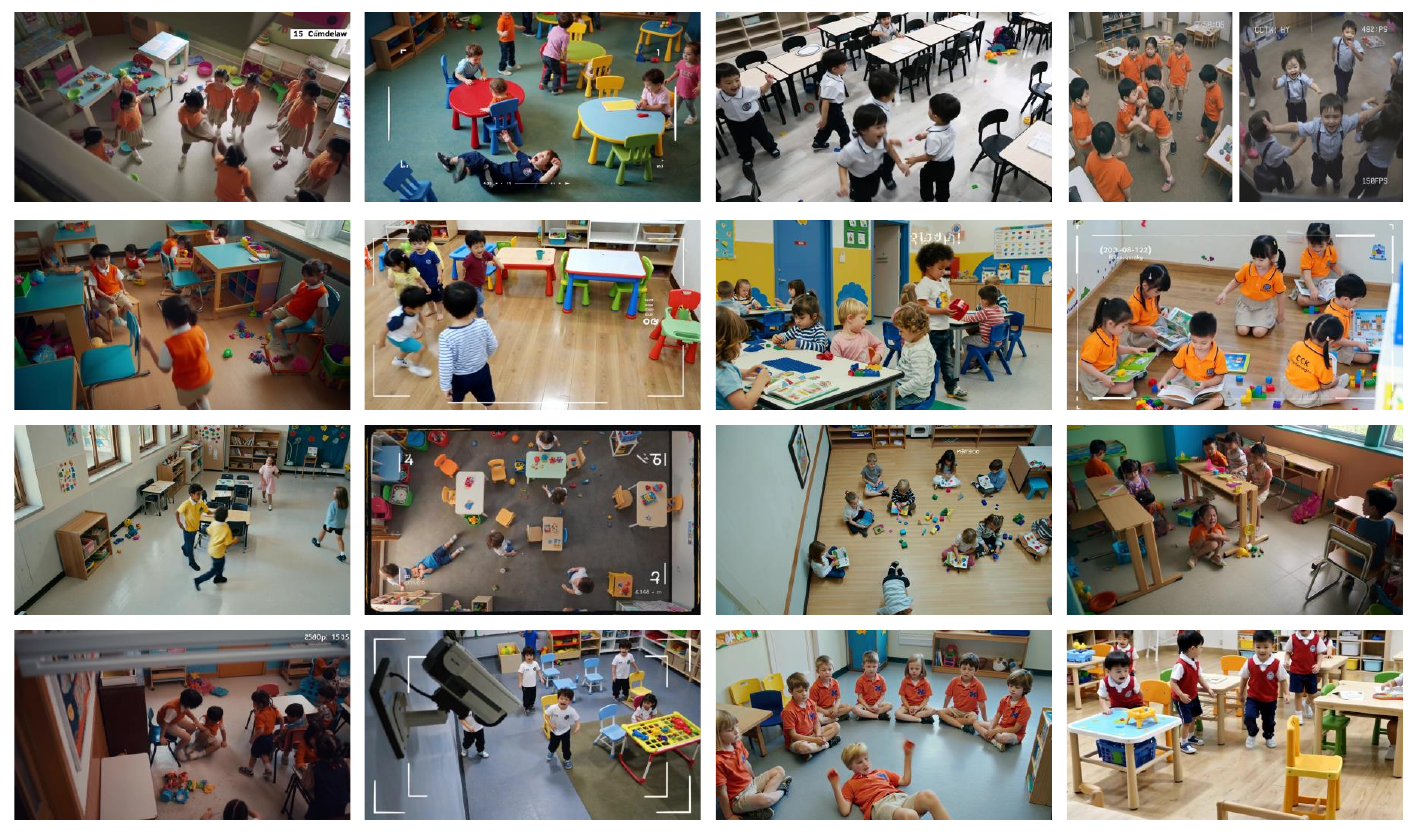}
   \caption{\textbf{Generated dataset samples.} Our synthetic dataset contains diverse viewpoints, lighting conditions, subjects, objects, classroom settings, clothing, etc. \textit{Please zoom in for better view.}}
   \label{fig:generated_dataset_samples}
\end{figure*}

\subsection{Annotation and Quality Check Process}

\paragraph{Annotation Tool Development.} 
We developed an integrated annotation toolbox for efficiently annotating human poses, action classes, and temporal boundaries for multiple human subjects in each scene. The tool improves annotation efficiency by providing a unified interface for different annotation types. It also preloads time-consuming annotations, such as skeletal poses, allowing annotators to correct inaccurate labels rather than annotating each video from scratch. This reduces the manual annotation burden and improves consistency across samples. We will release the annotation toolbox for use by other researchers.

\paragraph{Annotations.} 
After generating all videos, we annotated the action classes and their temporal boundaries using the proposed annotation toolbox.

\paragraph{Quality Checks by Preschool Teachers.} 
After annotation, all videos and labels were validated by preschool teachers, who served as domain experts. Samples rejected by the teachers were discarded from the dataset. This validation step is important because preschool teachers have direct experience observing classroom behavior and safety-related incidents in real educational settings. Their review helps ensure that the generated videos are visually plausible, that the depicted incidents are realistic, and that the action labels and temporal annotations accurately reflect the intended motion and action semantics. In total, we collect 1296 synthetic dataset samples. Examples from our final dataset are visualized in \autoref{fig:generated_dataset_samples}.

\paragraph{Ethics and Privacy Considerations.}
Since this work involves classroom environments and visual data of children, all data collection and handling procedures followed the approved institutional ethics protocol. All authors involved in data collection, handling, and annotation completed the required IRB/human-subjects research training before accessing any human-subject data. 
Real student images were anonymized and used only as references for generating non-identifiable uniform assets. 
The released dataset will contain only synthetic videos and approved real-world predicted skeleton keypoints, with no personally identifiable or linkable student or teacher information.

\subsection{Realworld Dataset}
In addition to creating synthetic videos, we also collected realworld data from actual preschools. Pose data belonging to each person in the classroom is extracted from anonymized video data. Action labels from previously mentioned selected action classes are annotated. We follow the same verification procedure as with synthetic dataset. Only verified annotations and anonymous pose data are retained and used for subsequent usage. In total, we collect 574 realworld samples.
\section{Approach}

\subsection{Overview}
Our approach is designed to recognize classroom safety-related incidents from pose sequences alone, reducing reliance on raw visual data. We assume that the recognition system has access only to human pose trajectories, rather than the underlying RGB video. In deployment, these trajectories may be extracted from classroom footage by the data-holding institution before being provided to the recognition model.

\begin{figure*}[t]
  \centering
  \includegraphics[width=0.8\linewidth]{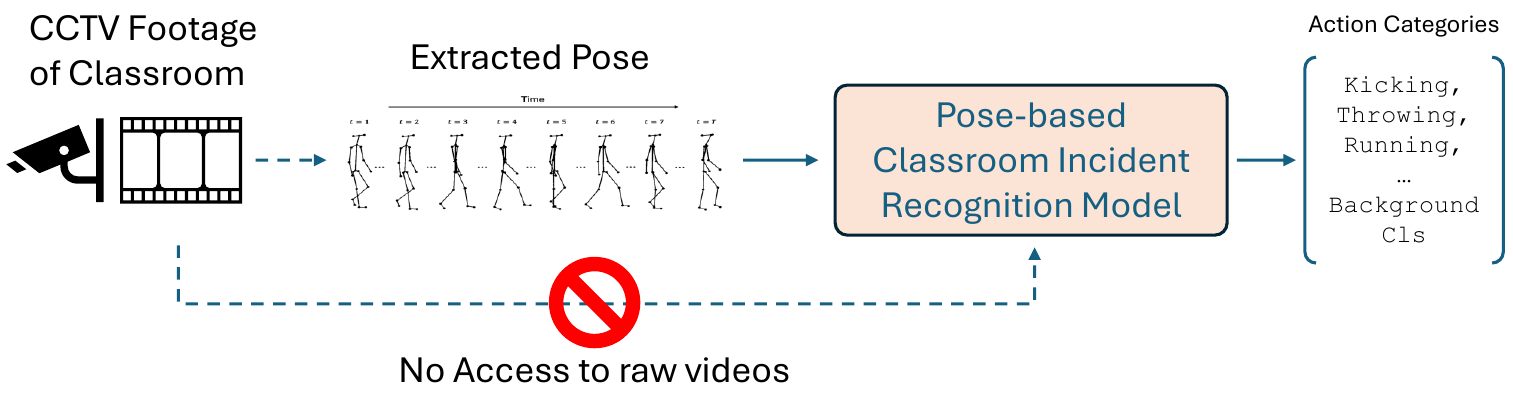}
   \caption{\textbf{Privacy-preserving pose-based classroom incident recognition approach.}}
   \label{fig:approach_overview}
\end{figure*}

Given pose trajectories, we first apply temporal preprocessing to mitigate pose-estimation noise. We then construct a hierarchy of kinematic representations, including joint positions, velocities, and accelerations. Each kinematic order is processed by a dedicated action-recognition backbone, and the resulting representations are fused to form a stronger hierarchical multi-order teacher  model. To reduce inference cost, we distill this fused teacher into a lightweight single-stream student model that operates only on zeroth-order joint-position features while retaining much of the teacher's predictive strength. Our full approach has been visualized in \autoref{fig:full_approach}.

\subsection{Pose-Based Incident Recognition}
Human action recognition can be performed directly from RGB video or from structured human representations such as skeleton and pose sequences. Compared with raw video, pose-based recognition reduces reliance on identifiable appearance information, including facial and other visual cues, making it a more privacy-conscious choice for classroom monitoring. Motivated by this consideration, we adopt pose-based incident recognition as the foundation of our framework. \autoref{fig:approach_overview} shows the overall pose-based privacy-preserving classroom incident recognition system.

\subsection{Hierarchical Kinematic Representations}
Joint trajectories are widely used in pose-based action recognition and have shown strong performance. However, position trajectories alone may not fully capture the dynamics of an action. Higher-order kinematic quantities, such as joint velocity and acceleration, can encode motion direction, speed, and changes in movement intensity that are not explicit in joint positions. These signals may be particularly useful for distinguishing classroom incidents involving rapid or forceful motion, such as punching, kicking, and throwing.

Motivated by this, we introduce a hierarchical kinematic representation that combines multiple motion orders. Specifically, we represent motion using joint positions, velocities, and accelerations, which are later encoded by dedicated order-specific backbones. Representations from later stages of these backbones are then fused to obtain a robust joint representation of human actions.

\paragraph{Computing higher-order kinematic features.}
We formalize the kinematic representations derived from pose sequences. Let \(\mathbf{j}_a^t \in \mathbb{R}^{C}\) denote the coordinate vector of joint \(a\) at time step \(t\), where \(C\) is the coordinate dimension. We treat joint positions as zeroth-order kinematic information. The corresponding first-order kinematic feature, i.e., velocity, is computed as

\begin{equation}
    \mathbf{v}_a^t =
    \frac{\mathbf{j}_a^t - \mathbf{j}_a^{t-\delta}}{\delta},
\end{equation}

where \(\delta\) denotes the temporal interval over which the displacement is measured. Similarly, the second-order kinematic feature, i.e., acceleration, is defined as

\begin{equation}
    \mathbf{a}_a^t =
    \frac{\mathbf{v}_a^t - \mathbf{v}_a^{t-\delta}}{\delta}.
\end{equation}

\paragraph{General form.}
More generally, the \(n\)-th-order kinematic quantity can be defined recursively as

\begin{equation}
    \mathbf{d}_a^{t,(n)} =
    \frac{
    \mathbf{d}_a^{t,(n-1)} -
    \mathbf{d}_a^{t-\delta,(n-1)}
    }{\delta},
\end{equation}

where \(\mathbf{d}_a^{t,(0)} = \mathbf{j}_a^t\) denotes the zeroth-order joint-position feature. In practice, higher-order kinematic features can therefore be constructed recursively from the pose sequence. Since these features can be precomputed once before training or inference, the associated computational overhead is minimal.

\begin{figure*}[t]
  \centering
  \includegraphics[width=0.8\linewidth]{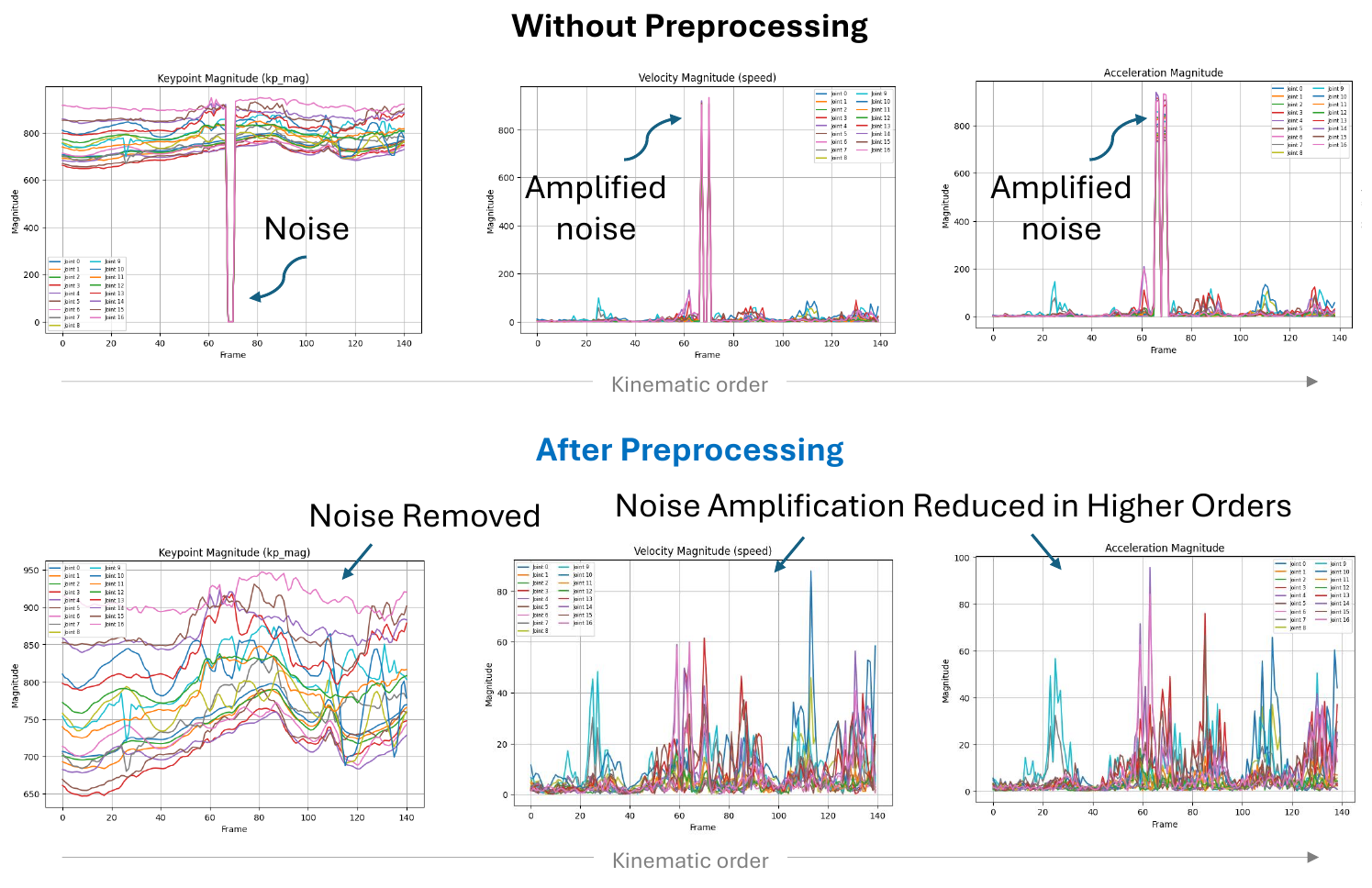}
   \caption{\textbf{Example Noise amplification when naively deriving higher order features.} Trajectories of three orders of kinematic data for all joints is plotted (different colors).}
   \label{fig:noise_amplification}
\end{figure*}

\paragraph{Pre-processing.} 
Higher-order kinematic features amplify errors present in lower-order pose estimates as shown in \autoref{fig:noise_amplification}. We therefore apply temporal preprocessing before computing derivatives. Missing joint coordinates are filled using linear interpolation over neighboring valid frames. To reduce frame-to-frame jitter and sudden pose-estimation artifacts, each joint trajectory is temporally smoothed using a Savitzky--Golay filter. Higher-order kinematic features, including velocities and accelerations, are then computed from the denoised trajectories.

\subsection{Unified Kinematic Action Recognition Model}
We define a unified action recognition model that can operate on any of the kinematic representations introduced above. For each person, frame-level joint features are aggregated over a temporal window to form a sequence of kinematic trajectories. These trajectories are then processed by an encoder backbone and a classification head to predict the corresponding action class.

Given an input sequence
\(
x \in \mathbb{R}^{T \times A \times C},
\)
where \(T\) denotes the number of time steps, \(A\) the number of joints, and \(C\) the feature dimension, the encoder backbone extracts a latent representation

\begin{equation}
    \phi = \theta_{BB}(x).
\end{equation}

The backbone is architecture-agnostic and may be instantiated using, for example, spatiotemporal graph convolutional networks, transformer-based models, or temporal convolutional networks. The resulting representation \(\phi\) is passed to a classification head to produce the predicted action class:

\begin{equation}
    \hat{y} = \theta_{H}(\phi),
\end{equation}

where \(\theta_H\) denotes the classification head. During training, both the encoder parameters \(\theta_{BB}\) and classifier parameters \(\theta_H\) are optimized jointly.

For notational simplicity, we do not specify the particular kinematic order used as input. The same model formulation applies to joint positions, velocities, accelerations, and other derived kinematic features, since they share the same overall trajectory structure, up to minor temporal padding differences introduced by derivative computation.

\paragraph{Order-specific backbones.}
We instantiate the previously described recognition model separately for each order of kinematic features. All backbones share the same architecture but maintain independent parameters. Each order-specific backbone is trained on its corresponding input representation. Specifically, the zeroth-order backbone is trained on joint-position trajectories, the first-order backbone on preprocessed velocity trajectories, and the second-order backbone on preprocessed acceleration trajectories. These backbones are pretrained independently, then frozen and fused as described next.

\subsection{Multi-Order Kinematic Fusion Approach}
We combine representations from different kinematic orders to obtain a unified multi-order action representation. As described above, each order-specific backbone is pretrained independently and then frozen. Given the latent representation \(\phi_m\) produced by the backbone associated with kinematic order \(m\), we fuse the order-specific features using weighted averaging:

\begin{equation}
\label{eq:phi_multiorder}
    \phi_{\text{multi-order}} =
    \frac{\sum_{m=0}^{M} w_m \phi_m}
    {\sum_{m=0}^{M} w_m},
\end{equation}

where \(M\) is the number of kinematic orders, \(\phi_m\) denotes the representation corresponding to order \(m\), and \(w_m\) is its associated fusion weight. Before fusion, representations from all order-specific backbones are projected into a common dimensional space. In our implementation, the weights \(w_m\) are set manually. The resulting fused representation \(\phi_{\text{multi-order}}\) serves as the final multi-order kinematic representation. It is then passed to a classification head to produce teacher predictions, which are subsequently used for distillation.

\begin{figure*}[t]
  \centering
  \includegraphics[width=0.9\linewidth]{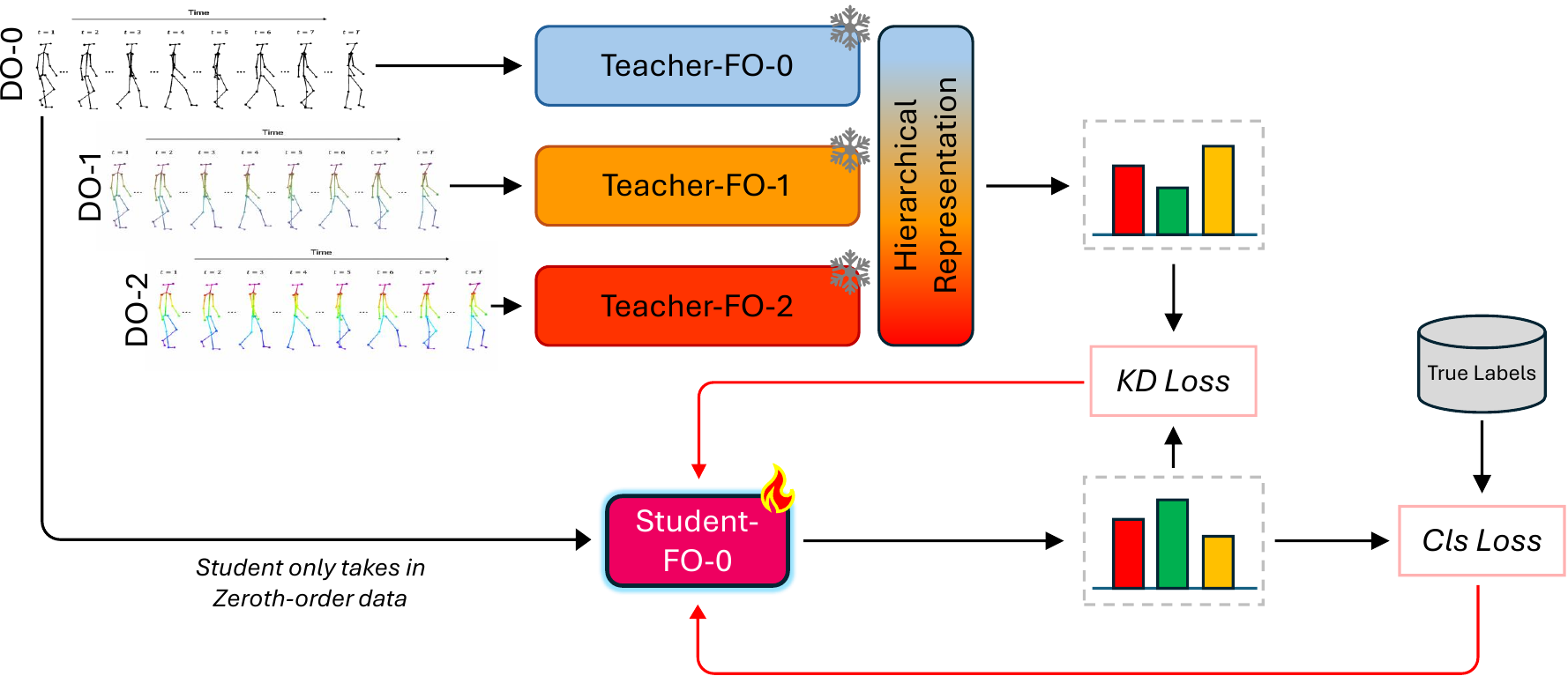}
   \caption{\textbf{Our full approach.} DO represents order of the input kinematic data; FO represents teacher feature's order.}
   \label{fig:full_approach}
\end{figure*}

\subsection{Multi-Objective Knowledge Distillation Approach}

Multi-order kinematic representations are expected to provide stronger and more robust action features. However, obtaining them requires running multiple order-specific backbones in parallel, which increases memory usage and computational cost. This limits their suitability for edge devices or other resource-constrained deployment settings. To address these bottlenecks, we introduce a distillation-based approach.

Specifically, we propose a multi-objective knowledge distillation (KD) framework that transfers information from the multi-order fused model into a substantially smaller zeroth-order student model. The student is trained to match the teacher’s softened class-probability distribution while operating only on a single zeroth-order feature stream. Thus, our framework performs compression along two axes: 1) from a teacher that leverages multi-order kinematic inputs to a student that uses only zeroth-order joint-position inputs; and 2) from a larger teacher network to a substantially smaller student network. We detail these two aspects below.

\paragraph{Distilling across feature orders.}
We distill knowledge from the multi-order kinematic model, which serves as the teacher, into a lighter zeroth-order student model. Since the student operates on only one kinematic input stream, it requires substantially less memory and computation than the multi-order teacher.

During training, the student is optimized using two complementary objectives. First, it is encouraged to match the softened class-probability distribution produced by the teacher, thereby transferring information about the teacher's class preferences and inter-class relationships. Second, it is trained with the ground-truth action labels to preserve direct supervision for the target recognition task.

Let \(z^{t}\) and \(z^{s}\) denote the teacher and student logits, respectively. Their temperature-scaled class-probability distributions are defined as

\begin{equation}
p_j^{t}(\tau) =
\frac{\exp(z_j^{t}/\tau)}
{\sum_k \exp(z_k^{t}/\tau)},
\qquad
p_j^{s}(\tau) =
\frac{\exp(z_j^{s}/\tau)}
{\sum_k \exp(z_k^{s}/\tau)},
\end{equation}

where \(\tau\) is the distillation temperature. The distillation loss is computed using the Kullback--Leibler divergence between the teacher and student output distributions:

\begin{equation}
\mathcal{L}_{KD}(p^{s}(\tau), p^{t}(\tau)) =
\tau^{2}\sum_{j}p_{j}^{t}(\tau)
\log\frac{p_{j}^{t}(\tau)}{p_{j}^{s}(\tau)}.
\end{equation}

In addition, the student is trained using a standard classification loss with the ground-truth action label \(y\):

\begin{equation}
\mathcal{L}_{Cls}(p^{s}(1), y) =
\sum_{j} -y_{j}\log p_{j}^{s}(1).
\end{equation}

The final multi-objective training loss is

\begin{equation}
\mathcal{L}_{MTL} =
\lambda_{KD}\mathcal{L}_{KD}
+
\lambda_{Cls}\mathcal{L}_{Cls},
\label{eq:loss_mtl}
\end{equation}

where \(\lambda_{KD}\) and \(\lambda_{Cls}\) control the relative contributions of distillation and supervised classification.

To summarize, the teacher learns richer motion cues from positions, velocities, and accelerations, but the final student keeps only a lightweight joint-position stream at inference. This gives the student some benefit of multi-order motion reasoning without paying the full computational cost.

\paragraph{Distilling into a smaller model.}
To further improve deployment efficiency, we not only compress the multi-order teacher into a zeroth-order student, but also design the student network to be substantially smaller than the teacher. Formally, we require

\begin{equation}
    |\theta_t| \gg |\theta_s|,
    \label{eq:small_model_condition}
\end{equation}

where \(|\theta_t|\) and \(|\theta_s|\) denote the numbers of trainable parameters in the teacher and student networks, respectively.
\section{Experiments}

\noindent\textbf{Implementation details.} We use PyTorch~\cite{pytorch} to implement all the models. Furthermore, we use PySKL toolbox~\cite{duan2022pyskl} in our work. We extract human pose from videos using YOLO pose estimator~\cite{redmon2016you, Jocher_Ultralytics_YOLO_2023}. We base our models on STGCN++ model~\cite{duan2022pyskl}, but our approach is not limited any particular model, and can be applied to various models. For our individual, order-specific teacher models, we adopt STGCN++ model with 64 base channels. For our student model, we adopt STGCN++, but with only 32 channels, reducing the parameters by half. For training our model, we use typical distillation temperature of 4; set $\{w_0, w_1, w_2\}$ to \{1.0, 0.8, 0.2\} and $\{\lambda_{KD}, \lambda_{Cls}\}$ to \{0.3, 1.0\}. For all models, we follow their recommended training criteria and pretrain models using NTURGBD large scale dataset~\cite{liu2019ntu}. For all models, we use 80\% of our synthetic dataset for training, while the remaining 20\% of the dataset serves as the test set. \\

\noindent\textbf{Performance metrics.} Following the prior literature, we use accuracy to measure and compare the performance of all the models.

\subsection{Comparison with Representative Methods on Synthetic Dataset}

\begin{table}
\small
    \centering
    \begin{tabular}{lc}
    \toprule
        \textbf{Model} & \textbf{Performance (\%)}\\
        \midrule
        AAGCN~\cite{aagcn} & 66.39\\
        STGCN++~\cite{duan2022pyskl} & 68.05\\
        CTRGCN~\cite{ctrgcn} & 65.56\\
        MSG3D~\cite{msg3d} & 68.88\\
        PoseC3D~\cite{posec3d} & 70.54\\
        Ours & \textbf{71.78}\\
        \bottomrule
    \end{tabular}
    \caption{\textbf{Performance comparison of various methods on our synthetic dataset in terms of accuracy (\%).}}
    \label{tab:synthetic_results}
\end{table}

We compare our method with various representative skeleton-based action recognition models, including AA-GCN~\cite{aagcn}, STGCN++~\cite{duan2022pyskl}, CTRGCN~\cite{ctrgcn}, MSG3D~\cite{msg3d} and PoseC3D~\cite{posec3d}. The results on our synthetic classroom incident dataset are reported in \autoref{tab:synthetic_results}.
Our method achieves the highest accuracy of 71.78\%, outperforming all compared baselines. In particular, it improves over PoseC3D, the strongest baseline on this dataset, by 1.24 percentage points, and over MSG3D by 2.90 percentage points. These results show that our framework is effective for recognizing classroom safety-related actions in the synthetic domain. Notably, our model is computationally efficient: it uses only one-tenth as many parameters as PoseC3D, yet achieves superior performance (\autoref{fig:accuracy-efficiency}).

The performance gain is consistent with our proposed multi-order motion reasoning strategy. While the compared skeleton-based baselines primarily learn from a single pose-based representation, our teacher model learns richer motion cues from multiple kinematic orders, including joint positions, velocities, and accelerations. Joint positions capture spatial pose configurations, velocities capture motion direction and speed, and accelerations capture changes in movement intensity. These cues are especially useful for recognizing rapid or forceful classroom incidents such as punching and kicking.
One possible explanation for the improved student performance is that distillation encourages the student to internalize motion-sensitive representations learned by the multi-order teacher. Although the deployed student uses only a lightweight zeroth-order joint-position stream at inference, joint-position trajectories still contain temporal structure from which velocity- and acceleration-like cues can be implicitly inferred. By matching the softened predictions of the fused multi-order teacher, the student may learn decision patterns that approximate the teacher's multi-order kinematic reasoning, even without explicitly computing or processing multiple kinematic streams at inference.
Thus, the improvement over representative baselines can be explained by the combination of denoised hierarchical kinematic representations, fused multi-order teacher learning, and efficient teacher-student distillation. This design allows the student to benefit from richer multi-order motion reasoning during training while retaining the low computational cost of a single-stream pose-based model during deployment.

Unlike many existing pose-based action recognition benchmarks, our dataset naturally combines challenges associated with both cross-subject and cross-view recognition. Therefore, although it contains fewer action classes than large-scale datasets, the performance of all evaluated methods remains comparatively lower. For instance, PoseC3D~\cite{posec3d} has been reported to achieve over 94\% accuracy on established benchmarks such as NTURGBD~\cite{liu2019ntu} and FineGym~\cite{shao2020finegym}, but reaches only 70.54\% accuracy on our classroom incident dataset. This performance gap suggests that classroom incident recognition presents distinct challenges, including viewpoint variation, subject diversity, noisy pose trajectories, and fine-grained differences between safety-related actions. While our method achieves the best performance among the compared approaches, the overall accuracy shows that the task remains far from solved. This leaves substantial room for future work on more robust pose-based models for privacy-preserving classroom safety monitoring.

\begin{table}
\small
    \centering
    \begin{tabular}{lc}
    \toprule
        \textbf{Model} & \textbf{Performance (\%)}\\
        \midrule
        AAGCN~\cite{aagcn} & 54.70\\
        STGCN++~\cite{duan2022pyskl} & 57.32\\
        CTRGCN~\cite{ctrgcn} & 53.83\\
        MSG3D~\cite{msg3d} & 59.23\\
        PoseC3D~\cite{posec3d} & 54.36\\
        Ours & \textbf{63.41}\\
        \bottomrule
    \end{tabular}
    \caption{\textbf{Performance comparison of various methods trained on our synthetic dataset and tested zero-shot on our real-world dataset.}}
    \label{tab:zero_shot_results}
\end{table}

\subsection{Zero-Shot Synthetic-to-Real Transfer}

To evaluate whether models trained on synthetic classroom incident videos can generalize to real-world classroom footage, we conduct a zero-shot synthetic-to-real transfer experiment. In this setting, models are trained only on the synthetic set and evaluated directly on the real-world test set, without fine-tuning on real-world samples. This protocol assesses the transferability of different representations under the domain shift between generated videos and real classroom recordings. Zero-shot transfer is more challenging than few-shot or multi-shot transfer as no target-domain training examples are used.

\begin{figure*}
    \centering
    \includegraphics[width=\linewidth]{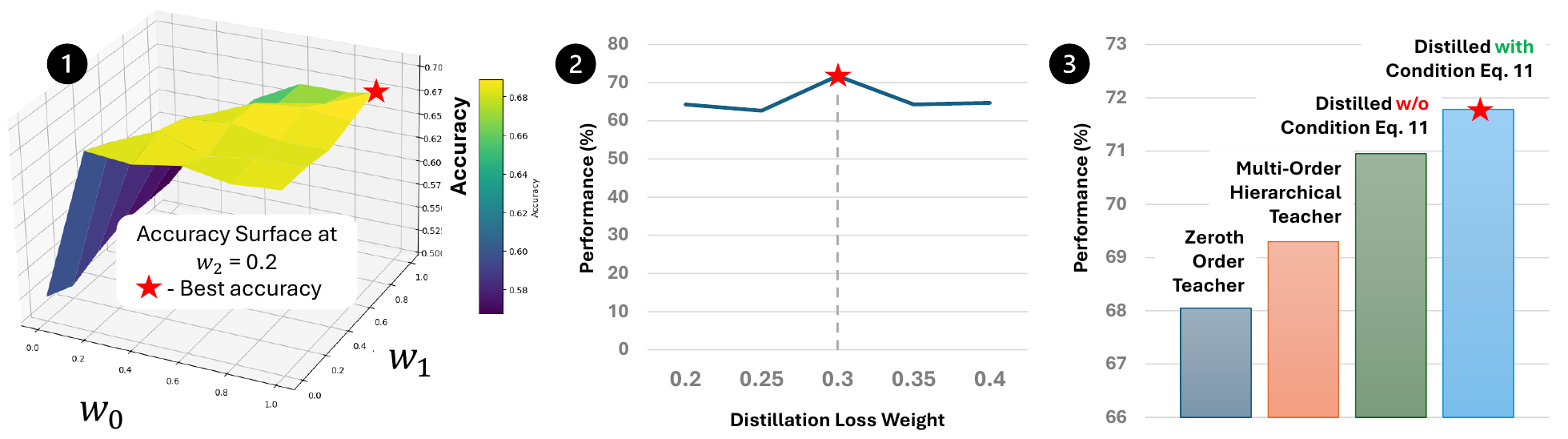}
    \caption{\textbf{Ablation study results.} (1) accuracy surface across a range of $w_0$ \& $w_1$; keeping $w_2$ constant at its optimal value of 0.2. (2) plot of accuracy across a range of distillation loss weight, $\lambda_{KD}$. (3) effect of multiorder hierarchical kinematic representation and distillation.}
    \label{fig:ablation_study_consolidated}
\end{figure*}

\autoref{tab:zero_shot_results} reports the zero-shot transfer results on the real-world dataset. All methods experience a performance drop when moving from synthetic to real-world data, confirming the presence of a domain gap. However, our method obtains the best real-world accuracy of 63.41\%, outperforming MSG3D, the strongest baseline in this setting, by 4.18 percentage points. Compared with the other baselines, the improvement is larger: 6.09 percentage points over STGCN++, 9.05 percentage points over PoseC3D, and 9.58 percentage points over CTRGCN.

The strong transfer performance of our method is likely due to its combination of pose-based abstraction, learned multi-order motion modeling, and multi-objective distillation. Pose features reduce dependence on synthetic visual appearance, while velocity and acceleration cues provide additional information about action dynamics. During training, the compact student model learns from both ground-truth labels and the richer soft predictions of the multi-order teacher, encouraging it to capture more informative and robust action representations. This helps the model preserve action-relevant motion patterns across the synthetic-to-real domain gap, explaining why it achieves the best zero-shot real-world accuracy while also requiring the fewest parameters and FLOPs at inference.

These results suggest that synthetic-domain accuracy alone is not sufficient to assess real-world usefulness. Although models such as PoseC3D perform strongly on the synthetic dataset, their accuracy decreases more substantially under zero-shot real-world evaluation. In contrast, our method achieves the best performance in both settings, indicating stronger cross-domain generalization under the synthetic-to-real shift.

At the same time, the decrease from 71.78\% on synthetic data to 63.41\% on real-world data shows that a synthetic-to-real domain gap remains. Therefore, these results should not be interpreted as showing that synthetic data fully replaces real-world data. Rather, they demonstrate that synthetic classroom incident videos can provide a useful training signal, and that the proposed method is comparatively more robust than existing baselines under zero-shot synthetic-to-real transfer.

\subsection{Ablation Study}

\paragraph{Ablating weights in multi-order kinematics fusion.} 
We study the effect of the contribution weights assigned to different kinematic orders in the hierarchical multi-order representation, as defined in \autoref{eq:phi_multiorder}. The results are shown in \autoref{fig:ablation_study_consolidated}~(1). We sweep each weight $w$ from 0 to 1 with a step size of 0.1 and record the resulting accuracy. The best performance is obtained with ${w_0, w_1, w_2}={1.0, 0.8, 0.2}$, indicating that all kinematic orders provide unique and complementary information.

\paragraph{Ablating loss weights in multitask loss.} 
We study the effect of the distillation loss weight $\lambda_{KD}$ in the multitask objective defined in \autoref{eq:loss_mtl}. Specifically, we train the model with different values of $\lambda_{KD}$ while keeping $\lambda_{Cls}=1.0$. The results are shown in \autoref{fig:ablation_study_consolidated}~(2). The best performance is obtained at $\lambda_{KD}=0.3$, suggesting that the classification and distillation losses provide complementary supervision.

\paragraph{Ablating multiobjective distillation.}
We further study the effects of hierarchical kinematic features and distillation conditions, with results shown in \autoref{fig:ablation_study_consolidated}~(3). First, hierarchical kinematic features improve over zeroth-order features, confirming the benefit of multi-order motion modeling. Second, we observe that distilling a multi-order hierarchical teacher into a single zeroth-order student allows the student to outperform the teacher. Third, imposing the condition in \autoref{eq:small_model_condition} further enables the student to outperform the much larger hierarchical teacher. We hypothesize that the student internalizes the teacher's multi-order motion reasoning while its smaller size (one sixth of the teacher size) encourages more effective and transferable representations.
Furthermore, compared with the base model, STGCN++~\cite{duan2022pyskl}, the proposed method improves accuracy by 4.1 percentage points and macro-F1 by 5.2 percentage points. As shown in the Appendix, these gains are mainly driven by better recognition of kick, sit, and run, as well as reduced confusion of kick and sit with the other class.
\section{Conclusion}
Classrooms are supposed to be safe places to learning and development. However, sometimes incidents take place that jeopardize the safety of children and teachers such as running, falling, violent activities such as punching or kicking. Advances in computer vision create strong potential for improving the safety of students and teachers in classroom environments via automated monitoring and incident recognition. However, this application has received comparatively little attention, resulting in limited study and a lack of resources to support both research and practical deployment. Thus, in this pilot study, we took a step toward bridging this gap. Firstly, we constructed a CCTV-based classroom incident recognition dataset using generative video AI, complemented by real-world data collection. Secondly, we developed a lightweight, privacy-preserving approach for recognizing classroom incidents. Our results show that this approach outperforms existing baselines while requiring fewer computational resources. We also conducted a novel zero-shot synthetic-to-real-world evaluation, in which our model outperformed all baselines. To encourage further progress in this area, we will publicly release our dataset, codebase, and supporting tools to the research community.

\paragraph{Acknowledgement.} This research/project is supported by Grant\# IHPC/SCC/G24-016.

{
    \small
    \bibliographystyle{ieeenat_fullname}
    \bibliography{main}
}

\clearpage



\end{document}